\documentclass{article}

\PassOptionsToPackage{numbers,square}{natbib}
\usepackage[main, final]{neurips_2026}

\usepackage[utf8]{inputenc} 
\usepackage[T1]{fontenc}    
\usepackage{hyperref}       
\usepackage{url}            
\usepackage{booktabs}       
\usepackage{amsfonts}       
\usepackage{amssymb}        
\usepackage{graphicx}       
\usepackage{nicefrac}       
\usepackage{microtype}      
\usepackage{xcolor}         
\usepackage{amssymb}
\usepackage{graphicx}
\usepackage{tikz}

\usepackage{amsmath}
\usepackage{multirow}
\usepackage{caption}
\usepackage[table]{xcolor}

\definecolor{first}{RGB}{255,190,190}
\definecolor{second}{RGB}{255,220,170}
\definecolor{third}{RGB}{255,245,170}

\title{VLSplat: Vision-Language Guided Object-Centric \\ 3D Gaussian Splatting via Scene Graph}

\author{
  Yuntae Jeon\thanks{Equal contribution.} \\
  Sungkyunkwan University \\
  \texttt{jyt0131@g.skku.edu}
  \And
  Younho Jeon\footnotemark[1] \\
  KAIST \\
  \texttt{117jyh@gmail.com}
  \And
  Sujin Jin \\
  Sungkyunkwan University \\
  \texttt{kane0113@skku.edu}
  \AND
  Sungho Jo\thanks{Corresponding authors.} \\
  KAIST \\
  \texttt{shjo@kaist.ac.kr}
  \And
  Seunghee Park\footnotemark[2] \\
  Sungkyunkwan University \\
  \texttt{shparkpc@skku.edu}
}

\begin{document}

\maketitle

\begin{abstract}

Learning semantic representations for 3D Gaussian Splatting has recently emerged as a promising direction for 3D scene understanding. 
However, existing 2D-to-3D semantic lifting approaches rely primarily on 2D segmentation supervision and lack object-level priors, causing the learned representation to be driven by local mask evidence rather than holistic object structure.
In this paper, we present \emph{VLSplat}, a vision-language guided approach for object-centric semantic lifting in 3D Gaussian Splatting. 
Our method augments mask-based lifting with object-level semantic priors derived from vision-language models and refines Gaussian representations using an object-group scene graph.
The refinement is performed through two language-guided modules: intra-group refinement suppresses semantic noise within object groups, while inter-group acquisition expands object support regions by incorporating structurally relevant candidates from neighboring groups. 
Extensive experiments on indoor scene datasets demonstrate that \emph{VLSplat} improves the semantic consistency and structural completeness of object-level categories.
Our code and models are available at \url{https://vlsplat.github.io/}.

\end{abstract}

\section{Introduction}
\label{1:introduction}

Understanding 3D scene semantics from multiple 2D views is a fundamental problem in computer vision, enabling object-level understanding and downstream applications such as robotics~\cite{zhang2025openvocabularyfunctional, jatavallabhula2023conceptfusion}, AR/VR, scene editing~\cite{zhuang2023dreameditor, chen2024gaussianeditor}, and semantic mesh extraction~\cite{chen2026sam, guedon2024sugar}. Recently, 3D Gaussian Splatting (3DGS)~\cite{Kerbl20233DGS} has emerged as a powerful representation for real-time novel-view synthesis, motivating growing interest in learning semantic-aware Gaussian representations for joint rendering and scene understanding. In this setting, the goal is to lift 2D semantic observations, typically segmentation masks, into object-aware 3D Gaussian representations such that 3DGS captures semantic properties of the underlying scene. 

Existing semantic 3DGS methods can be broadly divided into language-feature lifting and mask-based grouping. 
Language embedding approaches learn language latent fields or attach vision-language features to Gaussian primitives for open-vocabulary 3D understanding. 
Language latent-field methods~\cite{zhai2025panogs, qin2024langsplat, li20254d, shi2024language, zuo2025fmgs, zhou2024feature} distill CLIP~\cite{radford2021learning} features into a continuous 3D feature space for natural-language querying, whereas Gaussian-level embedding methods~\cite{li2025langsplatv2,wu2024opengaussian} associate such features with individual primitives for more discrete point-level semantics. 
However, these methods mainly lift or associate vision-language semantics with reconstructed Gaussians for retrieval and querying, rather than directly optimizing Gaussian features with vision-language guidance for category-aware object representation learning.

In contrast, mask-based grouping methods use 2D segmentation masks to learn Gaussian groups at instance-level, often producing more explicit object separation than language-field representations~\cite{ying2024omniseg3d, ye2024gaussian, lyu2024gaga, siddiqui2023panoptic}. 
Recently, codebook-based lifting has been introduced for 3DGS semantic lifting, where Unified-Lift~\cite{zhu2025rethinking} associates Gaussian features with a trainable object-level codebook to organize primitives into mask-consistent object regions.
However, such methods are induced by local 2D segment masks such as  SAM~\cite{kirillov2023segment}, whose view-dependent and multi-level predictions can split the same object into multiple parts or merge semantically different regions into one mask, especially in indoor scenes. 
As a result, the learned Gaussian groups may appear locally consistent in rendered views, but they are not explicitly grounded in category-level object semantics, limiting complete object-centric representation learning.

In this work, we propose VLSplat, a vision-language guided framework for category-aware object-centric semantic lifting in 3DGS. Our goal is to extend codebook-based Gaussian lifting from category-agnostic instance grouping to category-aware object representation learning under 3DGS optimization. We first construct hybrid 2D supervision by combining text-prompted object masks for major indoor categories with category-agnostic masks for the remaining regions. This provides semantic object cues while retaining dense mask coverage for the scene.

To overcome noise and inconsistency from 2D masks, we reinterpret learned Gaussian-codebook groups as nodes in a scene graph. 
Each node represents a semantic object candidate or an uncertain fragment, and graph relations provide spatial context for reasoning about whether a group is a meaningful object, noisy component, or candidate region to be reassigned. 
Based on this graph, VLSplat performs two complementary refinements: intra-group semantic refinement suppresses noisy primitives within object nodes, while inter-group acquisition expands incomplete object support by retrieving structurally plausible candidates from surrounding background regions using vision-language priors.

We conduct extensive experiments on 3D semantic and instance segmentation and point-level semantic segmentation. Following the thing--stuff taxonomy~\cite{narita2019panopticfusion}, VLSplat targets thing categories, i.e., objects with a well-defined shape such as chairs, desks, and beds.
Rather than merely lifting vision-language semantics onto reconstructed Gaussians for retrieval or querying, our method directly optimizes Gaussian features with vision-language guidance through scene-graph structured refinement.
Our contributions are summarized as follows:

\begin{itemize}
    \item We propose \emph{VLSplat}, a vision-language guided semantic lifting framework that extends codebook-based Gaussian grouping from category-agnostic instance clustering to object-centric representation learning.

    \item We design an object scene graph that interprets codebook groups as nodes enriched with VLM-derived language priors and spatial adjacency, and use it as an object-level prior for Gaussian refinement.
    
    \item We introduce language-guided Gaussian gating on synthetic object views, enabling intra-node refinement and inter-node acquisition that improve both the semantic consistency and structural completeness of object groups.
\end{itemize}

\section{Related Work}
\label{2:related_work}


\paragraph{2D-to-3D Semantic and Instance Lifting.}
2D-to-3D lifting transfers 2D masks, segments, or semantic features into 3D fields while enforcing multi-view consistency. Neural-field-based methods lift object selections, panoptic masks, or contrastive segments into volumetric representations~\cite{siddiqui2023panoptic, ren2022neuralvolumetric}, while related approaches incorporate open-vocabulary semantics, promptable segmentation, or hierarchical grouping into radiance fields~\cite{chen2024panoptic, cen2023segmentanything3d, kim2024garfield}. With the rise of 3DGS, later methods lift 2D mask supervision onto explicit Gaussian primitives for object-level instance segmentation~\cite{ying2024omniseg3d, ye2024gaussian, lyu2024gaga, dou2024cosseggaussians, choi2024clickgaussian, piekenbrinck2025opensplat3d}. 
Unified-Lift~\cite{zhu2025rethinking} further introduces codebook-based Gaussian semantic lifting, where Gaussian features are associated with a trainable object-level codebook to form mask-consistent object regions. Recent works also revisit the lifting solver itself through structured self-training, global label assignment, or modular segmentation on pre-trained 3DGS scenes~\cite{wang2025plgs, shen2024flashsplat, chacko2025lifting}. However, existing lifting methods remain strongly tied to 2D mask quality and cross-view consistency, whereas our proposed VLSplat injects object-level language priors and graph-guided reassignment to refine ambiguous or incomplete Gaussian groups.

\paragraph{Vision-Language Gaussian Embedding.}
Recent works extend 3DGS to language-aware scene understanding by embedding semantic, CLIP~\cite{radford2021learning}-aligned, or foundation-model features into Gaussian representations~\cite{qin2024langsplat, shi2024language, zuo2025fmgs, zhou2024feature, dou2024semanticgaussians}. 
These methods distill vision-language features into continuous 3D language fields or Gaussian primitives for open-vocabulary retrieval, natural-language querying, and rendered-view semantic prediction~\cite{qin2024langsplat, li20254d, shi2024language, zuo2025fmgs, zhou2024feature, qiu2024gls, chen2025ovgaussian}. 
Recent SAM~\cite{kirillov2023segment}+CLIP~\cite{radford2021learning} based approaches improve object-level localization by combining object masks with language features: OpenGaussian~\cite{wu2024opengaussian} and LangSplatV2~\cite{li2025langsplatv2} link SAM-derived instance features with CLIP embeddings for point-level open-vocabulary understanding, and PanoGS~\cite{zhai2025panogs} uses language-guided graph cuts with SAM-guided affinities for panoptic scene understanding. 
Beyond per-scene optimization, SceneSplat~\cite{li2025scenesplat} and Chorus~\cite{li2026chorus} pretrain generalizable 3DGS scene encoders with vision-language or multi-teacher supervision, following self-supervised point representation learning such as Sonata~\cite{wu2025sonata}. 
While effective for queryable semantic fields and transferable scene encoders, these methods mostly treat language features as annotations to be lifted or distilled, whereas VLSplat uses object-level language priors to structurally refine Gaussian object groups.

\paragraph{Structured Scene Graph Reasoning.}
3D scene graphs provide structured abstractions for objects, regions, agents, and their relations in 3D environments. 
Early works construct hierarchical scene graphs for indoor scenes and reconstructed environments, linking object-level semantics with indoor-, or scene-level structure~\cite{armeni20193dscenegraph, wald2020learning3dssg, wu2021scenegraphfusion}. 
Later mapping systems extend scene graphs to online spatial perception, dynamic agents, and robot planning through metric-semantic maps and layered graph structures~\cite{ros20213ddynamicscenegraphs, ros2021kimera, hughes2022hydra}. 
Recent language-aware scene graphs further support open-vocabulary scene abstraction, open-set relation prediction, language-guided navigation, and context-aware entity grounding~\cite{gu2024conceptgraphs, deng2024opengraph, werby2024hovsg, koch2024open3dsg, chang2023ovsg, koch2024lang3dsg}. 
While prior works mostly use scene graphs as output abstractions, structured memories, or downstream reasoning substrates~\cite{yan2025dovsg, saxena2025grapheqa, koch2025relationfield, chen2024clip}, VLSplat uses graph structure as an optimization scaffold for refining object-centric Gaussian groups during semantic lifting.

\section{Proposed Approach}
\label{3:proposed_approach}

\begin{figure}[t]
  \centering
  \includegraphics[width=\textwidth]{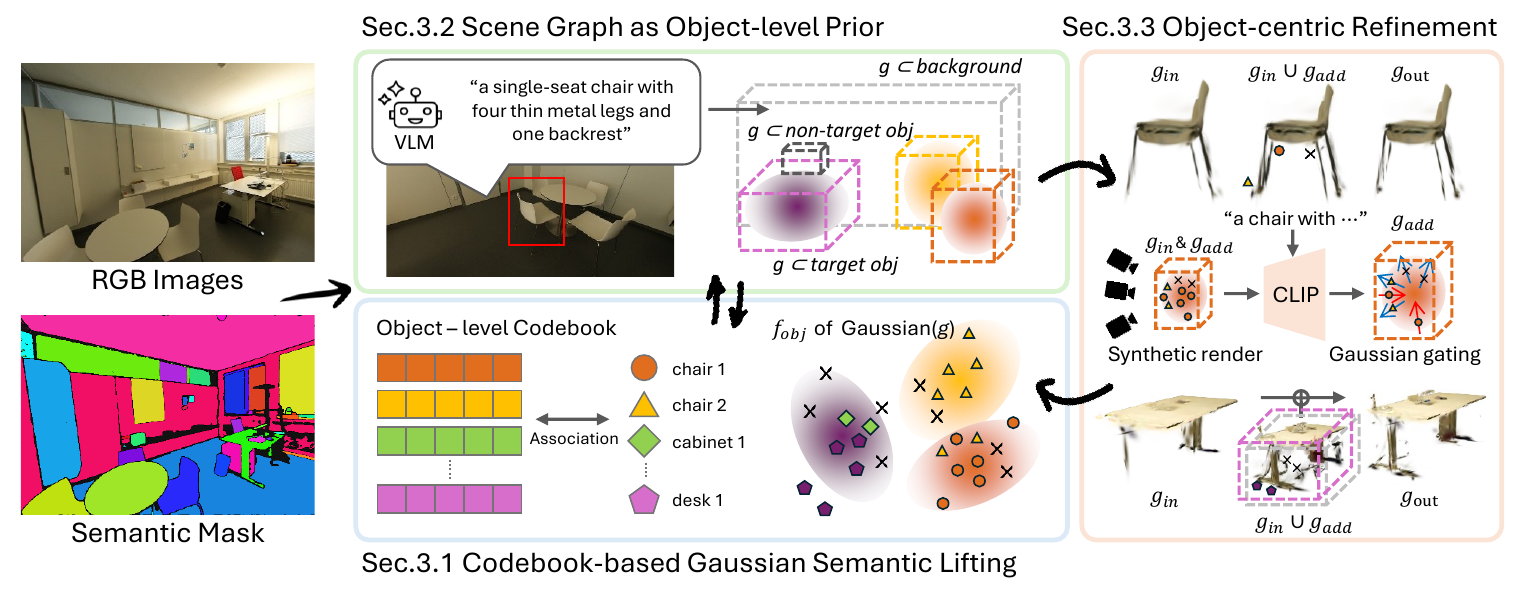}
  \captionsetup{justification=raggedright,singlelinecheck=false}
  \caption{
  \textbf{Overview of VLSplat.} 
  Starting from posed RGB images and object-aware semantic masks, our method learns Gaussian-level semantic features and associates them with an object-level codebook to form object groups. 
  Throughout 3D Gaussian optimization, the semantic representation and scene graph are iteratively updated, allowing object-centric reasoning over Gaussian groups. 
  The scene graph further guides language-based intra-group refinement and inter-group acquisition, improving semantic consistency and structural completeness.
  }
  \label{fig:main_vlsplat}
\end{figure}

VLSplat addresses object-level 3D instance and semantic segmentation in 3D Gaussian Splatting scenes, illustrated in Fig. ~\ref{fig:main_vlsplat}. 
The method first lifts object-aware 2D semantic masks into 3D with Gaussian-level features and an object-level codebook (Sec.~\ref{3:1}). 
It then constructs a 3D scene graph that associates codebook groups with language semantics and spatial context (Sec.~\ref{3:2}). 
Finally, VLSplat uses the scene graph as an object-level prior to refine Gaussian features beyond mask-level supervision, producing more object-consistent semantic representations (Sec.~\ref{3:3}).


\subsection{Codebook-based Gaussian Semantic Lifting}
\label{3:1}


\paragraph{Object-level codebook.}
Object-level 3D segmentation requires grouping dense Gaussian primitives into coherent object regions.
To this end, we employ a codebook-based lifting module that learns object-aware Gaussian features from 2D semantic supervision.
Given a 3D Gaussian scene \(\mathcal{G}=\{g_i\}_{i=1}^{N}\), each Gaussian \(g_i\) has a 3D center \(\boldsymbol{\mu}_i\), covariance \(\boldsymbol{\Sigma}_i\), opacity \(\alpha_i\), color attributes \(\mathbf{c}_i\), and a learnable object feature \(\mathbf{f}_i \in \mathbb{R}^{d}\).
We jointly learn the Gaussian object features and a compact object-level codebook \(\mathbf{F}_{\mathrm{obj}}=[\mathbf{p}_1,\mathbf{p}_2,\cdots,\mathbf{p}_K]^{\top}\in \mathbb{R}^{K \times d}\), where \(K\) is the maximum number of object-level groups and \(\mathbf{p}_k\) denotes the \(k\)-th latent group prototype.
Each Gaussian is assigned to the prototype with the highest dot-product association score, i.e., \(\hat{k}_i=\arg\max_k \mathrm{sim}(\mathbf{f}_i,\mathbf{p}_k)\), where \(\mathrm{sim}(\mathbf{f}_i,\mathbf{p}_k)=\mathbf{f}_i^{\top}\mathbf{p}_k\).

The quality of these initial groups depends heavily on the 2D semantic supervision used for lifting. In indoor scenes, generic mask supervision such as SAM2~\cite{ravi2024sam} often produces view-inconsistent masks or fragmented object regions across views, leading to unstable object grouping in 3D space. To reduce this ambiguity, we construct object-aware semantic maps using SAM3~\cite{carion2025sam} with category-level text prompts for major indoor objects, and use SAM2 to fill the remaining unlabeled regions, detailed in supplementary. 
This hybrid supervision combines category-aware masks with dense scene coverage, enabling the learned codebook to produce more stable coarse object hypotheses for subsequent scene-graph construction.

\subsection{Scene Graph as Object-level Prior}
\label{3:2}

Given the coarse object hypotheses from codebook lifting, we construct an ROI-based scene graph for object-centric refinement.
We assign language-aware evidence from training views and augment each node with 3D spatial support.
We further use a VLM to convert category-level evidence into object-specific descriptions, providing richer priors for subsequent representation learning. 


\paragraph{Spatial object support.}
Given the codebook assignment \(\hat{k}_i\), we collect each codebook group as \(G_k=\{i\mid \hat{k}_i=k\}\) and retain its high-confidence support \(S_k=\{i\in G_k \mid \langle \bar{\mathbf{f}}_i,\bar{\mathbf{p}}_k\rangle>\tau_{\mathrm{cb}}\}\). Although \(S_k\) provides a high-confidence codebook support, feature confidence alone does not ensure spatial coherence.
We therefore construct geometry-aware edges over \(S_k\) using center distance and anisotropic covariance.
For two Gaussians \(i,j\in S_k\), an edge is added when their projected supports overlap along the center-to-center direction:

\begin{equation}
(i,j) \in \mathcal{E}_k^{\mathrm{geo}}
\quad \mathrm{when} \quad
\|\boldsymbol{\mu}_j-\boldsymbol{\mu}_i\|_2
\leq
\alpha
\left(
\sqrt{\hat{\mathbf{d}}_{ij}^{\top}\boldsymbol{\Sigma}_i\hat{\mathbf{d}}_{ij}}
+
\sqrt{\hat{\mathbf{d}}_{ij}^{\top}\boldsymbol{\Sigma}_j\hat{\mathbf{d}}_{ij}}
\right),
\end{equation}
where \(\hat{\mathbf{d}}_{ij}\) is the normalized direction from \(\boldsymbol{\mu}_i\) to \(\boldsymbol{\mu}_j\), and \(\alpha\) controls the overlap tolerance. The connected components induced by \(\mathcal{E}_k^{\mathrm{geo}}\) define spatially coherent candidate supports.
After discarding spatially isolated components, we aggregate the remaining components as the spatial support of group \(k\), fit a 3D bounding box to obtain the tight support \(B_k^{\mathrm{tight}}\) for intra-node refinement, and expand it to derive \(B_k^{\mathrm{exp}}\) for inter-node refinement.

\paragraph{Language-prior object semantics.}

After obtaining the spatial support of each codebook group, we assign language-aware semantics and node types using multi-view evidence. Specifically, we project the Gaussian support into the training views and compare the resulting 2D regions with the object-aware semantic maps. We also use a VLM to help interpret the category-level evidence and generate an object-specific description. The resulting description captures visual and structural cues, including shape, parts, material, and contextual appearance, providing object-level language evidence for CLIP~\cite{radford2021learning}-based optimization. We provide an ablation study on the effect of VLM prompts on object representation learning in the supplementary material.

\paragraph{Scene graph construction.}
The final scene graph is defined as \(\mathcal{H}=(\mathcal{V},\mathcal{E})\), where each node \(v_k\in\mathcal{V}\) represents a semantic object candidate and stores the codebook ID \(k\), semantic category, VLM-generated object description, codebook prototype \(\mathbf{p}_k\), tight ROI \(B_k^{\mathrm{tight}}\), and expanded ROI \(B_k^{\mathrm{exp}}\).
Each node also retains SAM3-based target-object evidence from the training views, which indicates whether the ROI consistently corresponds to a meaningful foreground object.
The edge set \(\mathcal{E}\) encodes neighboring-object relations: two nodes \(u\) and \(v\) are connected when their expanded ROIs overlap in 3D, i.e., \(\mathrm{Vol}(B_u^{\mathrm{exp}}\cap B_v^{\mathrm{exp}})>0\).
This graph provides two complementary priors for object-centric refinement: \textbf{semantic grounding} for interpreting latent codebook groups as object nodes, and \textbf{spatial constraints} for restricting ambiguous Gaussian candidates to plausible neighboring object regions.

\subsection{Two-stage Object-centric Refinement}
\label{3:3}

\paragraph{Intra-node refinement.}
The first stage resolves semantic ambiguity within a target node \(v \in \mathcal{V}\) of the scene graph \(\mathcal{H}\).
Although Gaussians assigned to the corresponding codebook are expected to support the same semantic object node, their features may oscillate due to noisy masks, boundary ambiguity, or partial visibility.
For the target node \(v\), we collect its assigned Gaussians within the node support as \(\mathcal{S}_v=\{i \mid \hat{k}_i=k_v,\ \boldsymbol{\mu}_i\in B_v\}\), and split them into a reliable core and an uncertain candidate set according to feature confidence:

{\small
\begin{equation}
\mathcal{G}_{\mathrm{in}}^{\mathrm{intra}}
=
\{i\in\mathcal{S}_v \mid s_i^{(v)}\ge\tau_f\},
\qquad
\mathcal{G}_{\mathrm{add}}^{\mathrm{intra}}
=
\mathcal{S}_v\setminus\mathcal{G}_{\mathrm{in}}^{\mathrm{intra}},
\end{equation}
}

where \(s_i^{(v)}=\langle \bar{\mathbf{f}}_i,\bar{\mathbf{c}}_{k_v}\rangle\) measures the alignment between Gaussian \(g_i\) and the prototype of the target node, and \(\tau_f\) is a confidence threshold.
The reliable core provides stable evidence for the target object node, while the uncertain set is further verified before feature reinforcement or suppression.

\begin{figure}[t]
  \centering
  \includegraphics[width=1\linewidth]{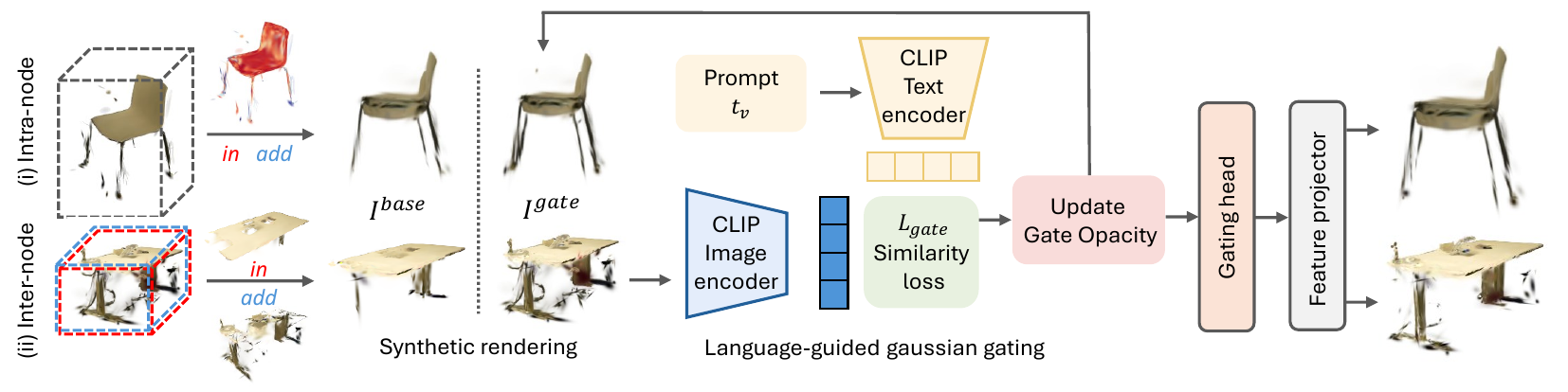}
  \caption{\textbf{Vision-language-guided object-centric refinement.}
The proposed two-stage refinement consists of \textbf{(i) intra-node refinement}, which resolves ambiguous Gaussians within a target node, and \textbf{(ii) inter-node acquisition}, which recovers missing object parts from neighboring non-target or background regions.
 }
\end{figure}

\paragraph{Inter-node acquisition.}
The second stage recovers missing parts of the target node \(v \in \mathcal{V}\).
Some Gaussians that spatially belong to the target object may be background or unlabeled groups because 2D supervision is incomplete or inconsistent across views.
For the target node \(v\), we construct an acquisition pool from all Gaussians inside the node support as \(\mathcal{T}_v=\{i \mid \boldsymbol{\mu}_i\in B_v\}\).
We then define the reliable core and acquisition candidates as

{\small
\begin{equation}
\mathcal{G}_{\mathrm{in}}^{\mathrm{inter}}
=
\{i\in\mathcal{T}_v \mid \hat{k}_i=k_v,\ s_i^{(v)}\ge\tau_f\},
\qquad
\mathcal{G}_{\mathrm{add}}^{\mathrm{inter}}
=
\{i\in\mathcal{T}_v \mid \hat{k}_i\notin\mathcal{X}_v\},
\end{equation}
}

where \(\mathcal{X}_v\) denotes the excluded foreground set, including the target node and neighboring foreground nodes that should not be directly absorbed.
Unlike intra-node refinement, which verifies uncertain Gaussians already assigned to the target node, inter-node acquisition tests whether non-target or background Gaussians inside the node support provide useful evidence for the target node.
This stage improves the semantic completeness of the target node without modifying the underlying geometry.

\paragraph{Synthetic object-view rendering.}
We evaluate candidate Gaussians \(\mathcal{G}_{\mathrm{add}}\) through object-centric rendering.
For each target node \(v\), we sample synthetic camera views around its 3D support and render the object from multiple directions.
To cover the full object extent, rendering is performed in an ROI-normalized local coordinate frame.
For each sampled view, we render a base image using only the reliable core and a gated image that additionally includes candidate Gaussians:

{\small
\begin{equation}
I_v^{\mathrm{base}}
=
R_v^{\mathrm{norm}}(\mathcal{G}_{\mathrm{in}}),
\qquad
I_v^{\mathrm{gate}}
=
R_v^{\mathrm{norm}}(\mathcal{G}_{\mathrm{in}}\cup q\odot\mathcal{G}_{\mathrm{add}}).
\end{equation}
}

where \(R_{v,m}^{\mathrm{norm}}\) centers and scales the local coordinate frame by the ROI of target node \(v\), allowing each sampled view to capture the full object extent.

\paragraph{Language-guided Gaussian gating.}
Each candidate Gaussian \(g_i\in\mathcal{G}_{\mathrm{add}}\) is assigned a learnable scalar gate \(q_i\in[0,1]\), which controls its opacity contribution during rendering.
The gate acts as an evidence estimator: a high value indicates that the candidate supports the target node prompt, while a low value indicates that the candidate is irrelevant or inconsistent. We optimize the gates using CLIP~\cite{radford2021learning} similarity between the rendered object and the node prompt \(t_v\).
Instead of maximizing the absolute CLIP score of the gated rendering, we maximize its improvement over the base rendering:
{\small
\begin{equation}
\mathcal{L}_{\mathrm{gate}}
=
-
\frac{1}{M}\sum_{m=1}^{M}
\left[
\operatorname{sim}(I_{v,m}^{\mathrm{gate}},t_v)
-
\operatorname{sim}(I_{v,m}^{\mathrm{base}},t_v)
\right],
\end{equation}
}

where $M$ is the number of synthetic views.
This relative objective is important because the base rendering already contains reliable object evidence; thus, the learned gates focus on candidates that provide additional semantic support beyond the current object core.

\paragraph{Object-level feature update.}
The optimized gates are used as semantic evidence to refine Gaussian object features without changing the geometry.
We update only the feature direction while preserving its original norm.
Let \(\mathbf{u}_i=\mathbf{f}_i/\|\mathbf{f}_i\|_2\) be the normalized feature direction of Gaussian \(g_i\). For each candidate \(g_i\in\mathcal{G}_{\mathrm{add}}\), we define a gate-dependent target direction 
\(\mathbf{t}_i=\bar{\mathbf{c}}_{k_v}\) if \(q_i\ge\tau_g\), and 
\(\mathbf{t}_i=-\bar{\mathbf{c}}_{k_v}\) otherwise, where \(\tau_g\) is the gate threshold.
The feature direction is then updated by directional interpolation:

{\small
\begin{equation}
\mathbf{u}_i'
=
\operatorname{norm}
\left(
(1-\rho_i)\mathbf{u}_i+\rho_i\mathbf{t}_i
\right),
\qquad
\mathbf{f}_i'
=
\|\mathbf{f}_i\|_2\mathbf{u}_i',
\end{equation}
}

where $\mathrm{norm}(\mathbf{x}) = \mathbf{x}/\|\mathbf{x}\|_2$ and $\rho_i$ is a per-Gaussian interpolation weight that moves the cosine similarity to the target prototype toward a bounded target value, with stronger repulsion for candidates with lower gate values.
Thus, high-gate candidates are attracted toward the target-node prototype direction, while low-gate candidates are repelled from it.
This updates semantic assignment cues while preserving Gaussian positions and scales.
Details on synthetic view sampling, refinement scheduling, and the handling of overlapping ROIs are provided in Sec.~\ref{sec:supp_impl} of the supplementary material.

\section{Experiments}


\begin{figure}[t]
  \centering
  \footnotesize
  \makebox[\linewidth][c]{%
    \begin{tabular}{@{}*{6}{c}@{}}
      \makebox[0.14\linewidth][c]{RGB} &
      \makebox[0.14\linewidth][c]{OpenGaussian} &
      \makebox[0.14\linewidth][c]{LangSplatV2} &
      \makebox[0.14\linewidth][c]{Unified-Lift + Ours} &
      \makebox[0.14\linewidth][c]{Ours} &
      \makebox[0.14\linewidth][c]{GT}
    \end{tabular}%
  }
  \vspace{2pt}
  \includegraphics[width=\linewidth]{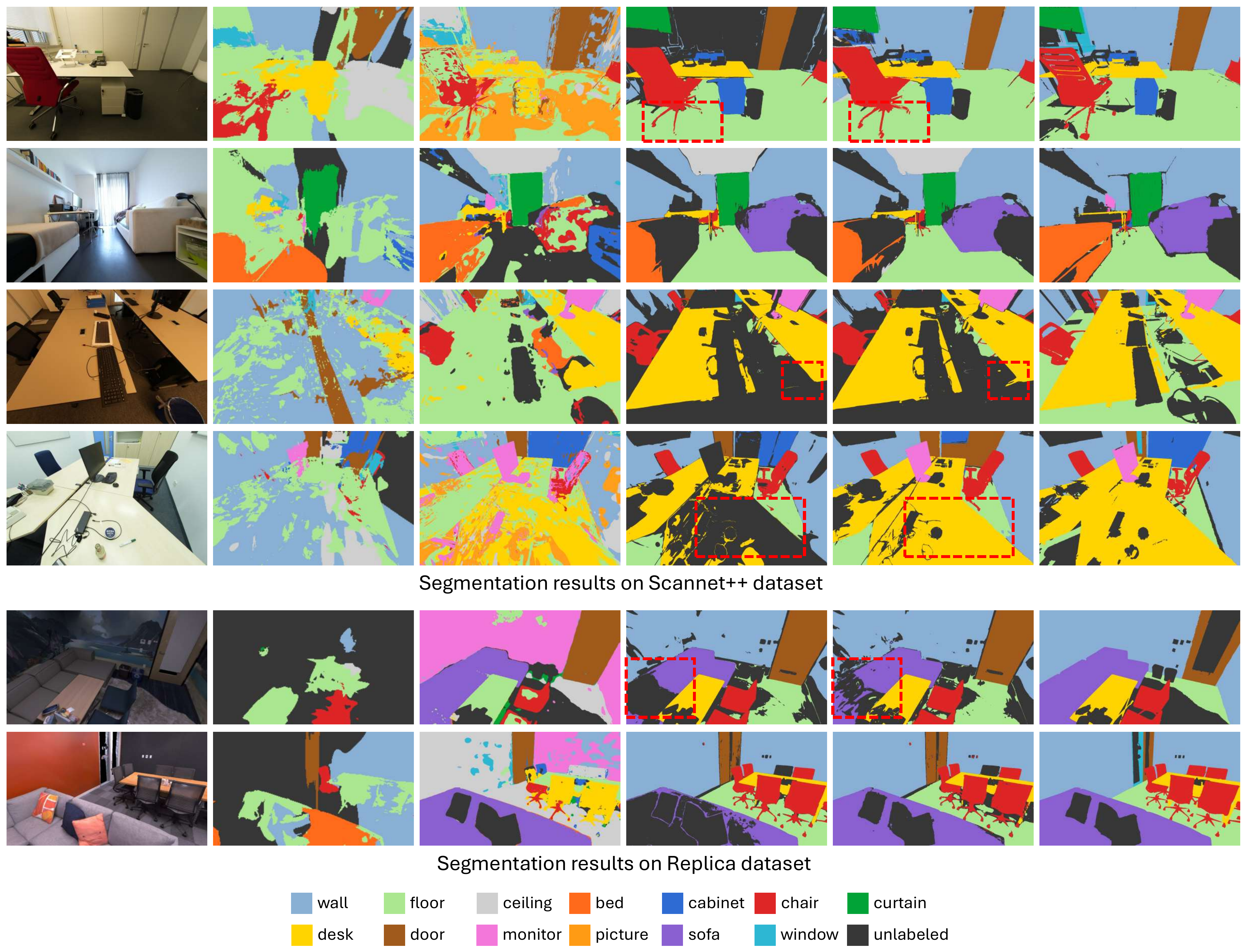}

  \caption{
  Qualitative render-level 3DGS semantic segmentation results on ScanNet++ and Replica.
  }
  \label{fig:qualitative_2d_segmentation}
\end{figure}


\begin{table}[b]
 \caption{Statistics of evaluation datasets used in our experiments. Each scene is optimized independently, and Train and Test denote the views of the same scene used for optimization and evaluation.}
  \label{tab:dataset_stats}
  \centering
  \small
  \setlength{\tabcolsep}{4pt}
  \renewcommand{\arraystretch}{0.75}
  \begin{tabular}{ll l l c ccc}
    \toprule
    \multirow{2}{*}{Protocol} &
    \multirow{2}{*}{Input Type} &
    \multirow{2}{*}{Dataset} &
    \multirow{2}{*}{Domain} &
    \multirow{2}{*}{Scenes} &
    \multicolumn{3}{c}{Average} \\
    \cmidrule(lr){6-8}
    & & & & & Objects & Train & Test \\
    \midrule
    \multirow{2}{*}{Render-level}
      & \multirow{2}{*}{RGB\&Sparse points}
      & ScanNet++~\cite{yeshwanthliu2023scannetpp}& DSLR       & 10 & 27 & 247 & 17 \\
      & 
      & Replica~\cite{straub2019replica}   & Simulation & 8 & 20 & 91 & 9 \\
    \midrule
    \multirow{2}{*}{Point-level}
      & \multirow{2}{*}{RGB\&Dense points}
      & ScanNetV2~\cite{dai2017scannet} & RGB-D      & 7 & 7 & 2965 & - \\
      &
      & Replica~\cite{straub2019replica}   & Simulation & 8 & 20 & 400 & - \\
    \bottomrule
  \end{tabular}
  \renewcommand{\arraystretch}{1.0}
\end{table}

\paragraph{Experimental setup.}
We evaluate our method on two complementary protocols summarized in Tab.~\ref{tab:dataset_stats}.
The first protocol follows the standard 3DGS optimization with pruning and densification. 
Since the learned representation is explicitly designed for differentiable rendering, evaluating rendered semantic maps allows us to measure the quality of the final 3DGS representation. 
The second protocol isolates the semantic lifting ability at the 3D point-level. 
For this evaluation, we freeze the input 3D primitives and disable pruning, densification, and position updates, following the point-level evaluation setting used in previous 3D Gaussian segmentation work~\cite{wu2024opengaussian}.

We adopt Unified-Lift~\cite{zhu2025rethinking} as the codebook-based lifting baseline and train it for 15k iterations to obtain coarse object-level Gaussian groups.
From 15k to 30k iterations, we update the scene graph every 500 iterations and perform object-level refinement based on the updated graph.
At each update, Gaussian gating is optimized for 40 iterations with mini-batches of four rendered views using the CLIP~\cite{radford2021learning} ViT-L model, after which intra-node refinement and inter-node acquisition are applied; full implementation details are provided in Sec.~\ref{sec:supp_impl} of the supplementary material.
All experiments, including semantic mask preparation, 3DGS training, scene-graph construction, and object-level refinement, are conducted on a single NVIDIA RTX 3090 GPU with 24GB VRAM. Results are reported from a single run per scene.

\paragraph{Thing--stuff taxonomy.}
Following the thing--stuff taxonomy in panoptic scene understanding~\cite{narita2019panopticfusion}, we report the metrics separately for our defined thing and stuff splits. Thing categories are bed, cabinet, chair, desk, monitor, and sofa, which serve as target nodes in our scene graph. Stuff categories are wall, floor, ceiling, curtain, door, picture, and window, which are evaluated during rendering but not used as primary targets for object-centric refinement.

\subsection{Render-level 3DGS Instance and Semantic Segmentation}

\paragraph{Datasets \& Evaluation metrics.}
We evaluate rendered 3D instance and semantic segmentation on ScanNet++~\cite{yeshwanthliu2023scannetpp} and Replica~\cite{straub2019replica}. 
We use 10 scenes from ScanNet++ and 8 scenes from Replica. 
In this protocol, models are trained in the standard 3DGS optimization setting, where densification, pruning, and Gaussian position updates are enabled. 
We evaluate semantic segmentation quality using mean accuracy (mAcc) and mean intersection-over-union (mIoU). 

We compare our method with OpenGaussian~\cite{wu2024opengaussian}, LangSplatV2~\cite{li2025langsplatv2}, and Unified-Lift~\cite{zhu2025rethinking}. 
Since the original Unified-Lift produces unlabeled object groups, ``Unified-Lift + Ours'' assigns category labels to its groups using our object-aware semantic maps and scene-graph based category association. It shares the same masks and training schedule as VLSplat but omits the iterative scene-graph refinement, allowing a direct comparison of the refinement under identical supervision.

\paragraph{Render-level 3DGS evaluation results.}
Tab.~\ref{tab:quantitative_segmentation} reports the quantitative results on ScanNet++ and Replica. 
VLSplat achieves strong performance on both datasets, with particularly consistent gains on Thing-level metrics. 
This demonstrates the effectiveness of our scene-graph-guided refinement for object-centric regions, where each target node can be grounded by category-aware masks, language descriptions, and local 3D support. 
For Stuff classes, the gains are smaller and in some cases the performance is slightly lower. 
This is consistent with the design of VLSplat, which focuses on foreground object nodes rather than large amorphous layout regions. 
The qualitative results in Fig.~\ref{fig:qualitative_2d_segmentation} further support this trend, showing that VLSplat produces more coherent object regions and better preserves fine-grained object parts.
Class-wise metrics are provided in the supplementary material.

\subsection{3D Point Cloud Semantic Segmentation}

\paragraph{Datasets \& Evaluation metrics.}
Following prior point-level 3D Gaussian segmentation works~\cite{wu2024opengaussian}, we evaluate 3D semantic and instance segmentation on ScanNetV2~\cite{dai2017scannet} and Replica~\cite{straub2019replica}.
Both datasets provide annotated 3D geometry with point-level semantic and instance labels, allowing predictions to be directly evaluated on fixed 3D support.
For semantic segmentation, we report mIoU and mAcc.
For instance-aware evaluation, we report panoptic reconstruction quality~\cite{dahnert2021panoptic}, i.e., PQ~\cite{kirillov2019panoptic} with an IoU matching threshold of 0.25, for thing and stuff categories, denoted as PRQ(T) and PRQ(S), respectively.

\begin{table*}[t]
  \caption{Quantitative 3DGS-based 2D render segmentation results on ScanNet++ and Replica.}
  \label{tab:quantitative_segmentation}
  \centering
  \small
  \setlength{\tabcolsep}{6pt}
  \begin{tabular}{lcccccccc}
    \toprule
    \multirow{3}{*}{Method}
    & \multicolumn{4}{c}{ScanNet++}
    & \multicolumn{4}{c}{Replica} \\
    \cmidrule(lr){2-5} \cmidrule(lr){6-9}
    & \multicolumn{2}{c}{mAcc}
    & \multicolumn{2}{c}{mIoU}
    & \multicolumn{2}{c}{mAcc}
    & \multicolumn{2}{c}{mIoU} \\
    \cmidrule(lr){2-3} \cmidrule(lr){4-5}
    \cmidrule(lr){6-7} \cmidrule(lr){8-9}
    & Thing & Stuff & Thing & Stuff
    & Thing & Stuff & Thing & Stuff \\
    \midrule
    LegGaussian~\cite{shi2024language}         & 27.09 & 33.73  & 20.89  & 13.19  & \multicolumn{4}{c}{(OOM*)} \\
    OpenGaussian~\cite{wu2024opengaussian}        & 19.09  & 37.69  & 18.33  & 26.10  & 12.89  & 9.32  & 12.03  & 7.46  \\
    LangSplatV2~\cite{li2025langsplatv2}         & \cellcolor{second}{65.69}  & \cellcolor{third}{57.82}  & \cellcolor{second}{61.32}  & \cellcolor{third}{39.94}  & \cellcolor{second}{66.88}  & \cellcolor{third}{56.40}  & \cellcolor{third}{59.64}  & \cellcolor{third}{35.80}  \\
    Unified-Lift~\cite{zhu2025rethinking} + Ours & \cellcolor{third}{64.77}  & \cellcolor{first}\bf{74.99}  & \cellcolor{third}{64.42}  & \cellcolor{first}\bf{68.63}  & \cellcolor{third}{65.99} & \cellcolor{second}{67.42} & \cellcolor{second}{65.93} & \cellcolor{second}{65.55} \\
    \midrule
    Ours                & \cellcolor{first}\bf{70.31}  & \cellcolor{second}{74.09}  & \cellcolor{first}\bf{69.81}  & \cellcolor{second}{67.79}  & \cellcolor{first}\bf{68.69} &  \cellcolor{first}\bf{68.11} &  \cellcolor{first}\bf{68.59} &  \cellcolor{first}\bf{67.75} \\
    \bottomrule
  \end{tabular}
\end{table*}

\begin{table}[t]
    \caption{Quantitative 3D point cloud segmentation results on ScanNetV2 and Replica. All methods are evaluated on the full scene under an identical protocol, except Ours~\textdagger, which restricts evaluation to scene graph-based ROIs to assess object-level quality.}
    
  \label{tab:pointcloud_segmentation}
  \centering
  \small
  \setlength{\tabcolsep}{4pt}
  \begin{tabular}{lcccccccc}
    \toprule
    \multirow{2}{*}{Method}
    & \multicolumn{4}{c}{ScanNetV2}
    & \multicolumn{4}{c}{Replica} \\
    \cmidrule(lr){2-5} \cmidrule(lr){6-9}
    & mIoU & mAcc & PRQ(T) $\uparrow$ & PRQ(S) $\uparrow$
    & mIoU & mAcc & PRQ(T) $\uparrow$ & PRQ(S) $\uparrow$\\
    \midrule
    OpenGaussian~\cite{wu2024opengaussian}        & 29.73 & 41.69 & - & - & 32.56 & 38.81 & - & - \\
    PanoGS~\cite{zhai2025panogs}              & \cellcolor{third}54.50 & \cellcolor{third}74.98 & \cellcolor{third}35.65 & \cellcolor{first}\bf{30.32} & 56.91 & 68.27 & 35.76 & \cellcolor{first}\bf{64.60} \\
    \midrule
    Ours                & 40.49 & 59.45 & 27.27 & 16.48 & \cellcolor{third}64.03 & \cellcolor{third}74.34 & \cellcolor{third}47.87 & 37.28 \\
    Ours \textdagger    & \cellcolor{first}\bf{61.89} & \cellcolor{first}\bf{79.12} & \cellcolor{first}\bf{46.18} & \cellcolor{third}27.80 & \cellcolor{first}\bf{72.20} & \cellcolor{first}\bf{80.37} & \cellcolor{first}\bf{54.80} & \cellcolor{third}38.21 \\
    \bottomrule
  \end{tabular}
\end{table}

\begin{figure}[t]
  \centering
  {\small
  \makebox[\linewidth][c]{%
    \makebox[0.19\linewidth][c]{RGB Reference}%
    \makebox[0.22\linewidth][c]{OpenGaussian}%
    \makebox[0.18\linewidth][c]{PanoGS}%
    \makebox[0.20\linewidth][c]{Ours}%
    \makebox[0.20\linewidth][c]{GT}%
  }}
  
  \vspace{2pt}
  \includegraphics[width=\linewidth]{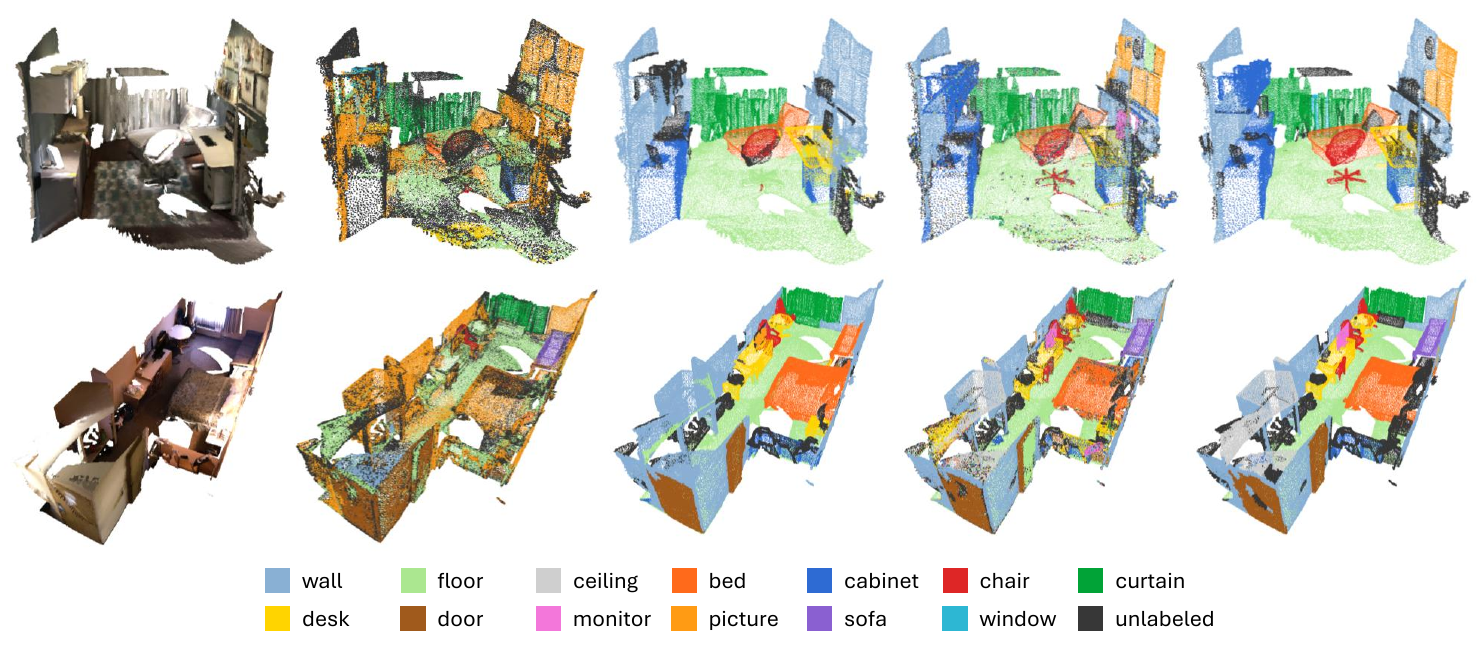}

  \caption{Qualitative 3D point-level segmentation results on ScanNetV2.}
  \label{fig:qualitative}
\end{figure}

\begin{table}[t]
  \centering
  \small
  \setlength{\tabcolsep}{2.2pt}
  \renewcommand{\arraystretch}{1.0}

  \begin{minipage}[t]{0.53\linewidth}
    \centering
    \caption{Scene graph construction ablation on \\ ScanNetV2 for object-level localization accuracy.}
    \label{tab:ablation_roi_construction}
    \begin{tabular}{c|ccc|cc}
      \toprule
      Case & Euc. & Aniso. & Feature & mIoU & mAcc \\
      \midrule
      \#1 &            &            &            & 40.49 & 59.45 \\
      \#2 & \checkmark &            &            & 42.70 & 63.64 \\
      \#3 &            & \checkmark &            & 43.88 & 64.91 \\
      \#4 &            &            & \checkmark & 58.51 & 76.11 \\
      \#5 &            & \checkmark & \checkmark & \bf{61.89} & \bf{79.12} \\
      \bottomrule
    \end{tabular}
  \end{minipage}
  \hfill
  \begin{minipage}[t]{0.44\linewidth}
    \centering
    \caption{Intra/inter-refinement ablation on \\   ScanNet++, Thing.}
    \label{tab:ablation_intra_inter}
    \begin{tabular}{c|cc|cc}
      \toprule
      Case & Inter & Intra & mAcc. & mIoU \\
      \midrule
      \#1 &            &            & 61.58 & 61.25 \\
      \#2 & \checkmark &            & 69.85 & 69.45 \\
      \#3 &            & \checkmark & 69.74 & 69.31 \\
      \#4 & \checkmark & \checkmark & \bf{70.31} & \bf{69.81} \\
      \bottomrule
    \end{tabular}
  \end{minipage}
\end{table}

\paragraph{Point-level 3DGS evaluation results.}
Tab.~\ref{tab:pointcloud_segmentation} reports point-level segmentation results on ScanNetV2 and Replica. Compared with OpenGaussian~\cite{wu2024opengaussian}, which assigns features to individual Gaussian primitives, our method benefits from organizing primitives into object-level nodes, leading to stronger semantic assignment on fixed 3D support. On Replica, VLSplat also outperforms PanoGS~\cite{zhai2025panogs} in both mIoU and mAcc, while PanoGS achieves higher full-scene scores on ScanNetV2. The two methods differ most at the object level (Fig.~\ref{fig:qualitative} and Fig.~\ref{fig:sup_3dpoint} in supp.). PanoGS embeds semantics in a continuous feature field, which tends to merge objects on a desk into the desk and to label the legs of chairs and desks as floor where they meet the ground. VLSplat keeps these as separate, complete segments by grouping Gaussians into discrete object nodes. This object-level quality is reflected in Ours~\textdagger, which evaluates within scene-graph ROIs and therefore complements, rather than directly compares with, the full-scene results. The relatively lower PRQ(S) suggests that our refinement mainly improves object-centric foreground nodes, while large background or layout regions are less directly refined.

\subsection{Ablation Study}

\paragraph{Ablation on scene graph.}
We analyze the effect of scene-graph ROI construction in Tab.~\ref{tab:ablation_roi_construction}. All variants are evaluated at the same training iteration, restricting evaluation to points within the resulting ROIs, while \#1 evaluates the full scene without ROIs. We compare Euclidean distance, covariance-aware anisotropic connectivity, feature similarity, and their combination for constructing object-level ROIs.
Feature similarity provides a large improvement over purely spatial grouping, indicating that semantic consistency is essential for localizing object nodes.
Combining feature similarity with anisotropic Gaussian connectivity achieves the best performance, showing that covariance-aware spatial support further improves ROI coherence and object-level localization.

\paragraph{Ablation on object-level refinement.}
We evaluate the contribution of intra-node refinement and inter-node acquisition in Tab.~\ref{tab:ablation_intra_inter}, with the scene graph construction fixed.
Each module alone substantially improves the baseline, and combining both achieves the best mAcc and mIoU, although the additional gain over either module alone is modest.
While the two modules operate on disjoint candidate sets, their effects overlap in practice: both mainly correct ambiguous Gaussians around object boundaries and thin structures under the same language prior, so once one module resolves the dominant ambiguities, fewer errors remain for the other.
The two modules nevertheless address complementary errors, as intra-node refinement removes incorrectly included Gaussians, whereas inter-node acquisition recovers missing object parts such as chair legs.

\section{Conclusion}

We presented \emph{VLSplat}, a vision-language guided framework for object-centric semantic lifting in 3D Gaussian Splatting. 
VLSplat addresses the limitations of mask-based lifting by using object-level language priors and scene-graph structure to refine Gaussian features beyond local 2D mask evidence. 
Through intra-node refinement and inter-node acquisition, our method suppresses semantic noise within object groups and recovers missing object support from neighboring regions. 
Experiments on render-level and point-level segmentation show that VLSplat improves semantic consistency and structural completeness, particularly for target object categories.

\paragraph{Limitations.}
Although VLSplat can recover missing object support caused by view-level mask inconsistency, it still requires a reasonable coarse object group to be formed during the initial lifting stage.
If an object is rarely observed or rarely detected across the input views, the codebook may fail to construct a reliable object node, and synthetic object-view rendering may provide insufficient visual evidence for CLIP-based gating.
Stronger 3D or generative priors, such as SAM3D~\cite{chen2026sam} or video diffusion models, could help complete such weakly observed regions, but they also introduce challenges in multi-view consistency and alignment with the reconstructed 3D coordinate system.

In addition, VLSplat relies on foundation models such as SAM3~\cite{carion2025sam}, a VLM, and CLIP~\cite{radford2021learning}, and its sensitivity to these choices has not been systematically studied. Our evaluation also focuses on thing categories in indoor scenes. Finally, the refinement introduces additional offline optimization cost that grows with the number of target nodes.

\paragraph{Broader Impact.}
VLSplat improves object-centric 3D scene understanding by organizing Gaussian primitives into semantically meaningful object representations. 
This can benefit downstream applications such as robotics, AR/VR, and semantic mapping. Such representations could support tasks requiring object-level
assessment of 3D scenes, such as progress monitoring in built
environments~\cite{jeon2025nerfbim}.
However, inaccurate semantic lifting may lead to incorrect object localization or unreliable scene interpretation, especially in safety-critical applications.

\newpage

\section*{Acknowledgments and Disclosure of Funding}

This research was supported by a National Research Foundation of Korea (NRF) grant funded by the Korean Government (MSIT) [RS-2024-00336270] and the National Research Foundation of Korea (NRF) grant funded by the Korea government (MSIT) [RS-2025–02223612].


{
\small
\bibliographystyle{unsrtnat}
\bibliography{references}
}

\newpage

\section*{Supplementary Material}

In this supplementary material, we provide the following:
\begin{itemize}
    \item \textbf{SAM2--SAM3 Mask Combination (Sec.~\ref{sec:a}):} Details of the hybrid 2D mask construction strategy, including mask sources, merging rules, post-processing, and qualitative comparisons.
    \item \textbf{Limitations of Prompt-based 2D Mask Generation (Sec.~\ref{sec:b}):} Analysis of view-inconsistent prompt-based masks and their effect on Gaussian semantic lifting.
    \item \textbf{Implementation Details of Scene Graph and Refinement (Sec.~\ref{sec:supp_impl})}: ROI expansion, synthetic view sampling, refinement scheduling, handling of overlapping ROIs, hyperparameters, and runtime.
    \item \textbf{Effect of VLM Prompts on Object-level Representation Learning (Sec.~\ref{sec:c}):} Qualitative analysis of object-level normalized rendering and CLIP-based Gaussian gating under different language prompts.
    \item \textbf{Additional Experiment Results (Sec.~\ref{sec:d}):} Additional quantitative and qualitative results, including per-class segmentation metrics, object-level rendering results, and detailed point-level segmentation visualizations.
\end{itemize}

\appendix

\section{SAM2--SAM3 Mask Combination}
\label{sec:a}

\paragraph{Mask sources.}
We use both SAM3~\cite{carion2025sam} and SAM2~\cite{ravi2024sam} to construct the 2D mask supervision used for Gaussian-level semantic lifting.
SAM3 is applied with category-level text prompts and provides semantically labeled masks for major indoor classes.
In our implementation, the prompt set includes wall, floor, ceiling, bed, cabinet, chair, curtain, desk, door, monitor, picture, sofa, and window.
These masks are useful for maintaining object-level semantic consistency, but they do not cover objects outside the prompt vocabulary and may miss small or weakly visible regions.

SAM2 is used through its automatic mask generator.
Unlike SAM3, SAM2 does not require text prompts and usually provides dense instance coverage over the image.
However, the resulting masks do not have category labels, and a single semantic object can be split into several small instances.
This behavior is problematic for our lifting stage because different parts of the same object may be assigned to different group IDs.

\paragraph{Compared variants.}
We consider three mask construction variants before applying the merged masks to Gaussian-level lifting.
The first variant is \textbf{SAM2-only}, where all masks are obtained from the SAM2 automatic mask generator.
This gives dense coverage, but often produces fragmented instances around object boundaries, furniture, and small cluttered regions.
The second variant is \textbf{SAM3-only}, where only text-prompted SAM3 masks are used.
This gives more stable semantic regions for prompted categories, but leaves uncovered pixels when objects are small or outside the predefined class set.
The third variant is our \textbf{SAM3-priority merged mask}, where SAM3 is treated as the main semantic source and SAM2 is used only to fill regions not sufficiently covered by SAM3.

\paragraph{Merging rule.}
Let $\mathcal{M}_{3}$ be the set of post-processed SAM3 masks and $\mathcal{M}_{2}$ be the set of SAM2 masks.
We first build a binary occupancy map from SAM3:
\begin{equation}
    O_{3}(x) = \bigvee_{m \in \mathcal{M}_{3}} m(x),
\end{equation}
where $x$ denotes a pixel location.
For each SAM2 mask $m_i \in \mathcal{M}_{2}$, we compute the fraction of its area that overlaps with the SAM3 occupancy map:
\begin{equation}
    r_i = \frac{|m_i \cap O_{3}|}{|m_i|}.
\end{equation}
A SAM2 mask is retained only when it is large enough and is not already explained by SAM3:
\begin{equation}
    |m_i| \geq 200
    \quad \mathrm{and} \quad
    r_i \leq 0.3.
\end{equation}
Masks smaller than 200 pixels are removed as small fragments.
Masks whose overlap ratio with SAM3 is larger than 0.3 are also removed, since they usually correspond to regions already covered by a semantically labeled SAM3 mask.
The final mask set is therefore
\begin{equation}
    \mathcal{M}_{\mathrm{merge}}
    =
    \mathcal{M}_{3}
    \cup
    \{m_i \in \mathcal{M}_{2} \mid |m_i| \geq 200,\ r_i \leq 0.3\}.
\end{equation}

\paragraph{SAM3 post-processing.}
Before merging, SAM3 masks are filtered to reduce duplicate or noisy semantic regions.
Small SAM3 masks with area below 500 pixels are removed.
We also suppress duplicate masks across related classes using mask overlap.
This is especially needed for categories such as \texttt{desk} and \texttt{table}, where the same physical object can be detected by both prompts.
After this step, SAM3 masks serve as the semantic backbone for the merged supervision.

\paragraph{Repainting.}
After filtering, the retained SAM2 masks are combined with the post-processed SAM3 masks and converted into a single 16-bit instance map.
When multiple masks overlap, masks are painted in ascending order of confidence score, so that higher-confidence masks overwrite lower-confidence ones.
This produces a single valid instance ID for each pixel and avoids ambiguous pixel assignments during lifting.

\begin{figure}[t]
    \centering
    \includegraphics[width=\linewidth]{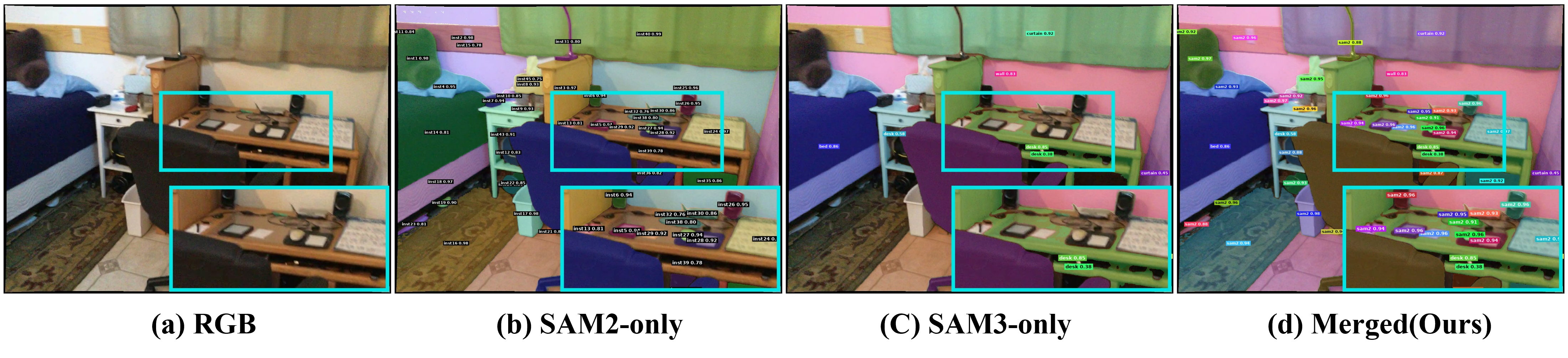}
    \caption{
    Qualitative comparison of mask construction variants.
    SAM2-only produces dense but fragmented instance masks.
    SAM3-only gives cleaner semantic masks for prompted categories but leaves several small regions uncovered.
    The merged result keeps SAM3 as the semantic backbone and adds SAM2 masks only in insufficiently covered regions.
    Cyan boxes show the enlarged desk region used for closer comparison.
    }
    \label{fig:sam2_sam3_merge}
\end{figure}

\paragraph{Qualitative comparison.}
Fig.~\ref{fig:sam2_sam3_merge} shows the difference between the mask variants.
SAM2-only covers most visible regions, but the desk area contains many small fragmented instances.
SAM3-only produces cleaner category-level regions for large objects such as the wall, curtain, bed, chair, and desk, but several small objects and local regions on the desk remain uncovered.
The merged result keeps the main semantic regions from SAM3 while adding SAM2 masks in the regions missed by SAM3.
This provides denser supervision without fully replacing the object-level consistency of SAM3.

\section{Limitations of prompt-based 2D mask generation.}
\label{sec:b}

\begin{figure}[htbp]
    \centering
    \includegraphics[width=\linewidth]{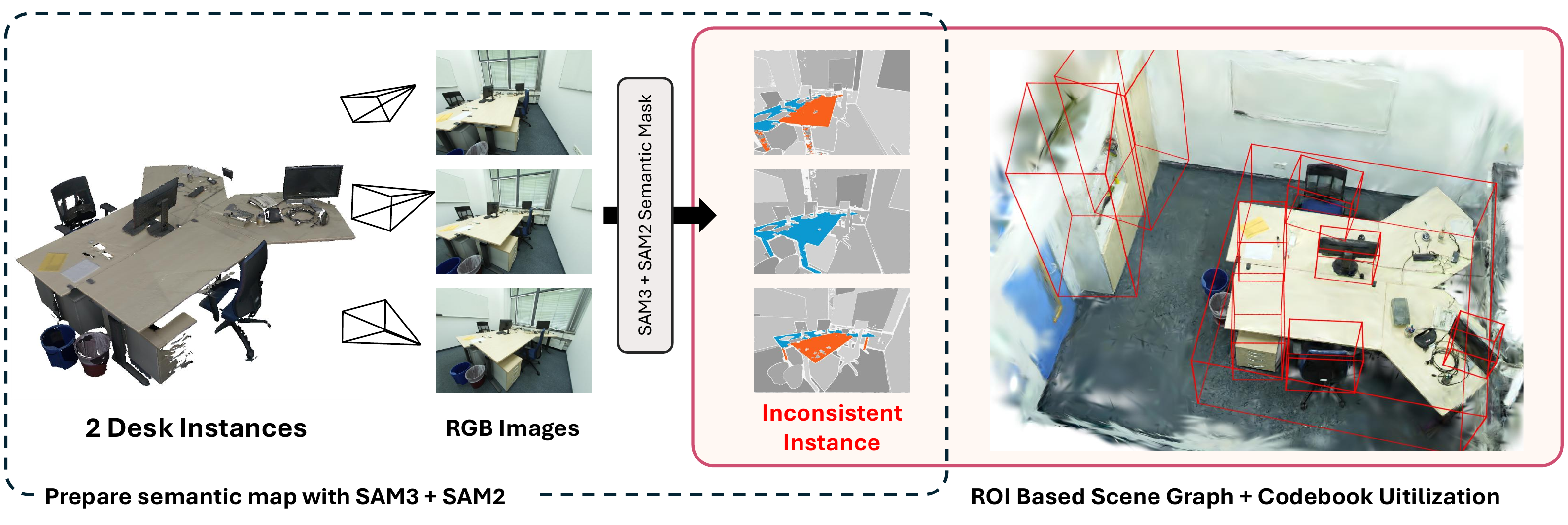}
    \caption{Failure cases of SAM3 under category-level prompting. Despite using the same ``desk'' prompt across consecutive views, SAM3 produces inconsistent instance-level segmentation results, alternately merging adjacent desks into a single region or separating them into distinct instances.
    }
    \label{fig:sup_inconsistent}
\end{figure}

Even with our SAM3~\cite{carion2025sam}+SAM2~\cite{ravi2024sam} mask generation strategy, prompt-based 2D segmentation can remain unstable across views. This issue is particularly common in indoor scenes, where objects of the same category are spatially adjacent or partially occluded. As shown in Fig.~\ref{fig:sup_inconsistent}, neighboring desks may be merged into a single region in one view while being separated into distinct instances in another. Such view-inconsistent masks provide ambiguous supervision for semantic lifting and can lead to fragmented or incorrectly merged Gaussian groups. These failure cases motivate the need for our ROI-based scene graph, which improves grouping consistency by reasoning over object relationships directly in 3D space.

\newpage

\section{Implementation Details of Scene Graph and Refinement}
\label{sec:supp_impl}

All settings below are shared across datasets without per-scene tuning, and the full configuration is released with the code.

\paragraph{ROI expansion.}
The expanded ROI $B^{\text{exp}}_k$ enlarges the tight ROI $B^{\text{tight}}_k$ by 10\% of its side length on both faces of every axis with the center fixed. Both ROIs are refit at every graph update.

\paragraph{Synthetic view sampling.}
Cameras orbit the ROI-normalized object center with azimuth in $[-180^\circ, 180^\circ]$, elevation in $[-30^\circ, 30^\circ]$, and radius in $[0.8, 1.2]$, using a $50^\circ$ field of view and $224\times224$ resolution. Each gate-optimization iteration uses $M=4$ views, two sampled views and their azimuth-opposite counterparts, and the CLIP improvement in Eq.~(5) is averaged over these views. Each graph update runs 40 iterations, and gates are thresholded at $\tau_g=0.7$.

\paragraph{Refinement schedule and overlapping ROIs.}
In each graph update, intra-node refinement is applied to all target nodes, followed by inter-node acquisition. Feature updates are deferred to the end of each update, so all nodes read the same features and the result does not depend on node order. When a Gaussian lies in the ROIs of multiple nodes, their proposed feature directions are averaged into a single update. 

\paragraph{Runtime and memory.}
Tab.~\ref{tab:supp_runtime} reports the cost of each stage on a single NVIDIA RTX 3090. For a scene with about 300 images and 10 target nodes, the full pipeline takes about 146--162 min, roughly $1.5$--$1.6\times$ the lifting baseline, with CLIP gating accounting for nearly all of the added cost. Peak memory increases by about 4 GB. At inference, VLSplat adds no graph, CLIP, or VLM computation, so rendering speed matches standard 3DGS.

\begin{table}[h]
\centering
\small
\caption{Runtime of each stage.}
\label{tab:supp_runtime}
\begin{tabular}{lll}
\toprule
Stage & Measured rate & Frequency \\
\midrule
SAM3 + SAM2 mask preparation & 1.8 s / image & Once per view \\
Codebook lifting & 3 min / 1k iterations & 3DGS optimization \\
Node definition & 1--2 min / scene & Once \\
VLM description generation & 3--4 s / node & Once \\
ROI computation & 1--2 s / node & Every graph update \\
Intra-node CLIP gating & 4--5 s / node & Every graph update \\
Inter-node CLIP gating & 4--5 s / node & Every graph update \\
\bottomrule
\end{tabular}
\end{table}

\section{Effect of VLM Prompts on Object-level Representation Learning}
\label{sec:c}

\begin{figure}
    \centering
    \includegraphics[width=1\textwidth]{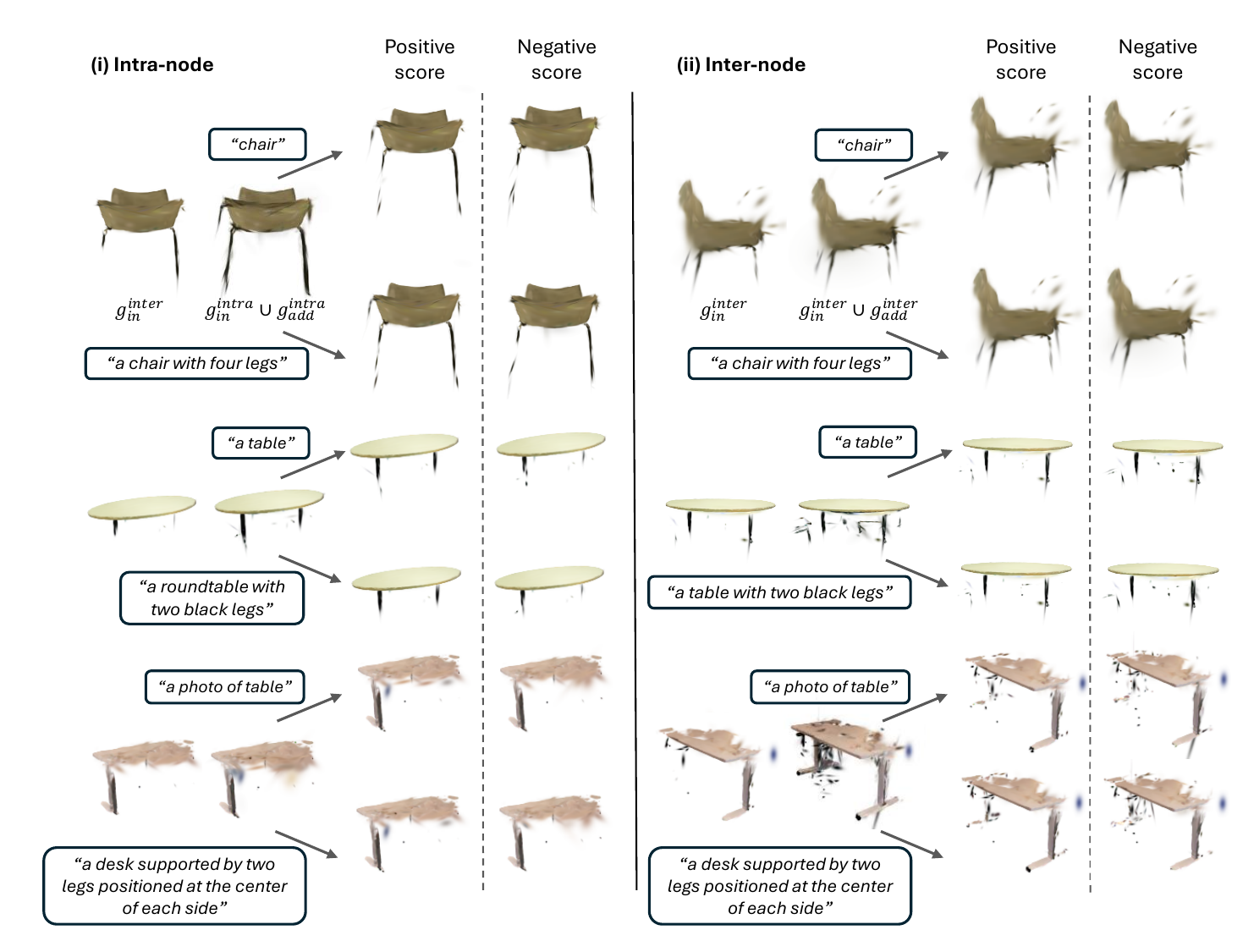}
    \caption{
    Object-level normalized view rendering and CLIP-based Gaussian gating results on ScanNet++.
    The first row shows a chair object from scene ``8b5caf3398'', the second row shows a table object from scene ``1ae9e5d2a6'', and the last row shows a desk object from scene ``67d702f2e8''.
    }
    \label{fig:sup_refinement}
\end{figure}

Fig.~\ref{fig:sup_refinement} visualizes the proposed intra-node and inter-node refinement process for object-centric Gaussian representation learning.
Following Sec.~\ref{3:2}, we first construct an object scene graph and identify the reliable object Gaussians $\mathcal{G}_{\mathrm{in}}$ and the candidate Gaussians $\mathcal{G}_{\mathrm{add}}$.
We then learn the opacity gates of the candidate Gaussians using CLIP~\cite{radford2021learning}-based language guidance.
Through this process, intra-node refinement suppresses noisy Gaussians inside the object region while encouraging object-consistent Gaussians to be treated as positive evidence, leading to more discriminative object features.
In contrast, inter-node refinement acquires candidate Gaussians from background or unlabeled groups when they are semantically and structurally consistent with the target object.

The results show that the proposed gating mechanism helps recover missing object parts while preventing irrelevant Gaussians from being incorporated.
In the first and third rows, inter-node refinement successfully includes thin structural parts such as chair or desk legs, which are often weakly observed in 2D masks or initially assigned to background Gaussian groups.
In the second row, although valid table-leg candidates are not sufficiently included in $\mathcal{G}_{\mathrm{add}}$, the learned gate still prevents noisy candidates from being incorrectly merged into the target object.
This indicates that the refinement process is useful not only for acquiring missing object parts but also for filtering unreliable candidates.

We further compare different text prompts used for CLIP-based Gaussian gating.
The prompts are constructed from object labels obtained from the scene graph and SAM3 masks, and are compared with more general prompt templates such as ``a photo of a \{class\}''.
Overall, more descriptive prompts that include color or structural attributes provide stronger language guidance for selecting appropriate Gaussians.
For example, in the first row, the prompt containing the phrase ``four legs'' assigns higher positive scores to Gaussians corresponding to the chair legs.
Similarly, in the table and desk examples, structure-aware prompts encourage the gating module to assign higher scores to Gaussians that better match the target object structure.
These qualitative results suggest that detailed VLM prompts improve CLIP-based Gaussian gating and contribute to more complete object-centric representation learning.

\section{Additional Experiment Results}
\label{sec:d}

\paragraph{Quantitative results on class-wise render-level 3DGS semantic segmentation}

\begin{table*}[t]
  \caption{Per-class quantitative segmentation results on ScanNet++ focus-object classes.}
  \label{tab:per_class_segmentation_scannetpp}
  \centering
  \scriptsize
  \setlength{\tabcolsep}{3.5pt}
  \begin{tabular}{lcccccccccccc}
    \toprule
    \multirow{3}{*}{Method}
    & \multicolumn{12}{c}{ScanNet++} \\
    \cmidrule(lr){2-13}
    & \multicolumn{6}{c}{Acc}
    & \multicolumn{6}{c}{IoU} \\
    \cmidrule(lr){2-7} \cmidrule(lr){8-13}
    & Bed & Cabinet & Chair & Desk & Monitor & Sofa
    & Bed & Cabinet & Chair & Desk & Monitor & Sofa \\
    \midrule
    LegGaussian
    & 43.61
    & 0.12
    & 5.96
    & \cellcolor{third}{71.09}
    & 0.00
    & 41.78
    & 37.62
    & 0.12
    & 5.96
    & 40.16
    & 0.00
    & 41.51 \\

    OpenGaussian
    & 9.69
    & 18.91
    & 29.50
    & 23.57
    & 17.74
    & 15.12
    & 8.99
    & 18.65
    & 26.77
    & 23.04
    & 17.46
    & 15.05 \\

    LangSplatV2
    & \cellcolor{third}{54.22}
    & \cellcolor{third}{51.10}
    & \cellcolor{third}{70.11}
    & 62.90
    & \cellcolor{first}\textbf{78.79}
    & \cellcolor{third}{77.02}
    & \cellcolor{third}{48.63}
    & \cellcolor{third}{50.73}
    & \cellcolor{third}{65.32}
    & \cellcolor{third}{58.61}
    & \cellcolor{first}\textbf{71.90}
    & \cellcolor{third}{72.73} \\

    Unified-Lift
    & \cellcolor{second}{59.08}
    & \cellcolor{second}{55.06}
    & \cellcolor{first}\textbf{77.10}
    & \cellcolor{second}{80.77}
    & \cellcolor{third}{26.29}
    & \cellcolor{second}{90.32}
    & \cellcolor{second}{59.06}
    & \cellcolor{second}{54.99}
    & \cellcolor{first}\textbf{76.55}
    & \cellcolor{first}\textbf{79.54}
    & \cellcolor{third}{26.21}
    & \cellcolor{second}{90.21} \\
    \midrule
    Ours
    & \cellcolor{first}\textbf{62.48}
    & \cellcolor{first}\textbf{59.78}
    & \cellcolor{first}\textbf{77.47}
    & \cellcolor{first}\textbf{81.26}
    & \cellcolor{second}{50.00}
    & \cellcolor{first}\textbf{90.88}
    & \cellcolor{first}\textbf{62.45}
    & \cellcolor{first}\textbf{59.62}
    & \cellcolor{first}\textbf{76.88}
    & \cellcolor{first}\textbf{79.59}
    & \cellcolor{second}{49.58}
    & \cellcolor{first}\textbf{90.73} \\
    \bottomrule
  \end{tabular}
\end{table*}

\begin{table*}[t]
  \caption{Per-class quantitative segmentation results on Replica focus-object classes.}
  \label{tab:per_class_segmentation_replica}
  \centering
  \scriptsize
  \setlength{\tabcolsep}{3.5pt}
  \begin{tabular}{lcccccccccccc}
    \toprule
    \multirow{3}{*}{Method}
    & \multicolumn{12}{c}{Replica} \\
    \cmidrule(lr){2-13}
    & \multicolumn{6}{c}{Acc}
    & \multicolumn{6}{c}{IoU} \\
    \cmidrule(lr){2-7} \cmidrule(lr){8-13}
    & Bed & Cabinet & Chair & Desk & Monitor & Sofa
    & Bed & Cabinet & Chair & Desk & Monitor & Sofa \\
    \midrule
    OpenGaussian
    & \cellcolor{third}{6.45}
    & \cellcolor{second}{41.07}
    & 19.75
    & 4.99
    & 0.00
    & 5.07
    & \cellcolor{third}{4.25}
    & \cellcolor{second}{38.48}
    & 19.40
    & 4.99
    & 0.00
    & 5.07 \\

    LangSplatV2
    & 5.03
    & \cellcolor{first}\textbf{94.04}
    & \cellcolor{third}{48.91}
    & \cellcolor{third}{62.77}
    & \cellcolor{first}\textbf{98.06}
    & \cellcolor{first}\textbf{92.46}
    & 3.29
    & \cellcolor{first}\textbf{71.27}
    & \cellcolor{third}{48.31}
    & \cellcolor{third}{60.50}
    & \cellcolor{first}\textbf{84.19}
    & \cellcolor{first}\textbf{90.26} \\

    Unified-Lift
    & \cellcolor{first}\textbf{56.91}
    & 32.99
    & \cellcolor{first}\textbf{89.81}
    & \cellcolor{first}\textbf{89.96}
    & \cellcolor{second}{73.94}
    & \cellcolor{third}{52.35}
    & \cellcolor{first}\textbf{56.91}
    & 32.96
    & \cellcolor{first}\textbf{89.63}
    & \cellcolor{first}\textbf{89.82}
    & \cellcolor{second}{73.94}
    & \cellcolor{third}{52.34} \\
    \midrule
    Ours
    & \cellcolor{first}\textbf{56.50}
    & \cellcolor{third}{33.77}
    & \cellcolor{first}\textbf{89.88}
    & \cellcolor{first}\textbf{89.59}
    & \cellcolor{third}{70.16}
    & \cellcolor{second}{72.24}
    & \cellcolor{first}\textbf{56.50}
    & \cellcolor{third}{33.73}
    & \cellcolor{first}\textbf{89.62}
    & \cellcolor{first}\textbf{89.37}
    & \cellcolor{third}{70.16}
    & \cellcolor{second}{72.16} \\
    \bottomrule
  \end{tabular}
\end{table*}

Tab.~\ref{tab:per_class_segmentation_scannetpp} reports per-class IoU and accuracy results for Thing categories, including bed, cabinet, chair, desk, monitor, and sofa. On ScanNet++, our method achieves the best performance for most categories. Results on Replica are additionally reported in Tab.~\ref{tab:per_class_segmentation_replica}. Although previous approaches also perform competitively on Replica due to its clean indoor scenes, our method performs on par with Unified-Lift on chair and desk, which are among the most dominant objects in indoor environments, and substantially improves sofa. Consequently, our method achieves the best overall Thing-category performance.

\begin{figure}
    \centering
    \includegraphics[width=1\textwidth]{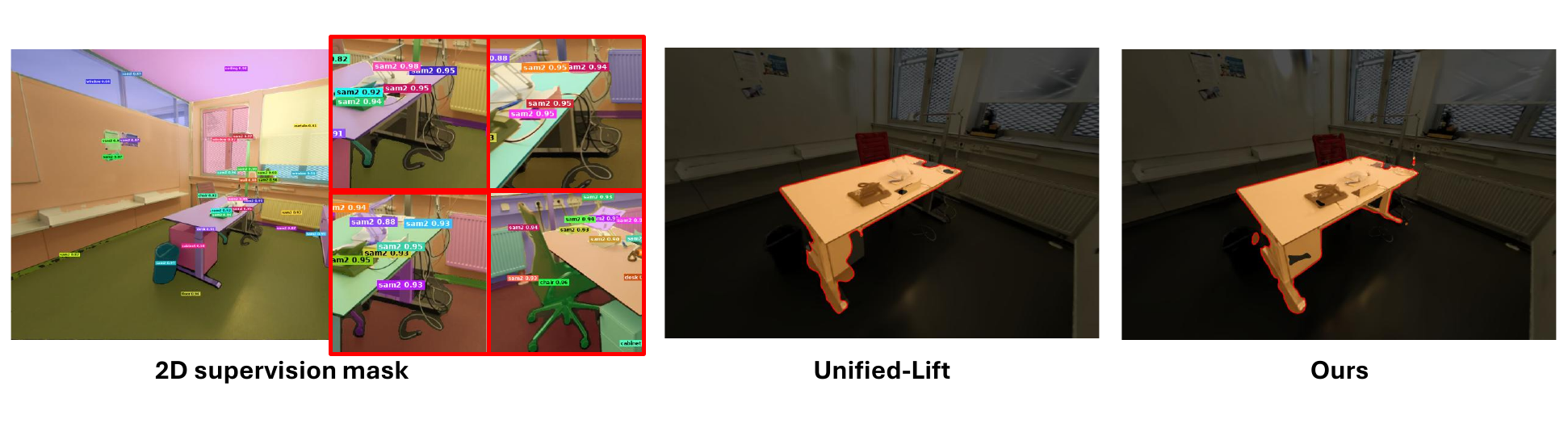}
    \caption{Qualitative results on language-guided object-centric representation learning from inconsistent 2D supervision on Scannet++ scene.
    }
    \label{fig:sup_qualitative2}
\end{figure}

\paragraph{Qualitative results on render-level 3DGS segmentation}
\begin{figure}
    \centering
    \includegraphics[width=1\textwidth]{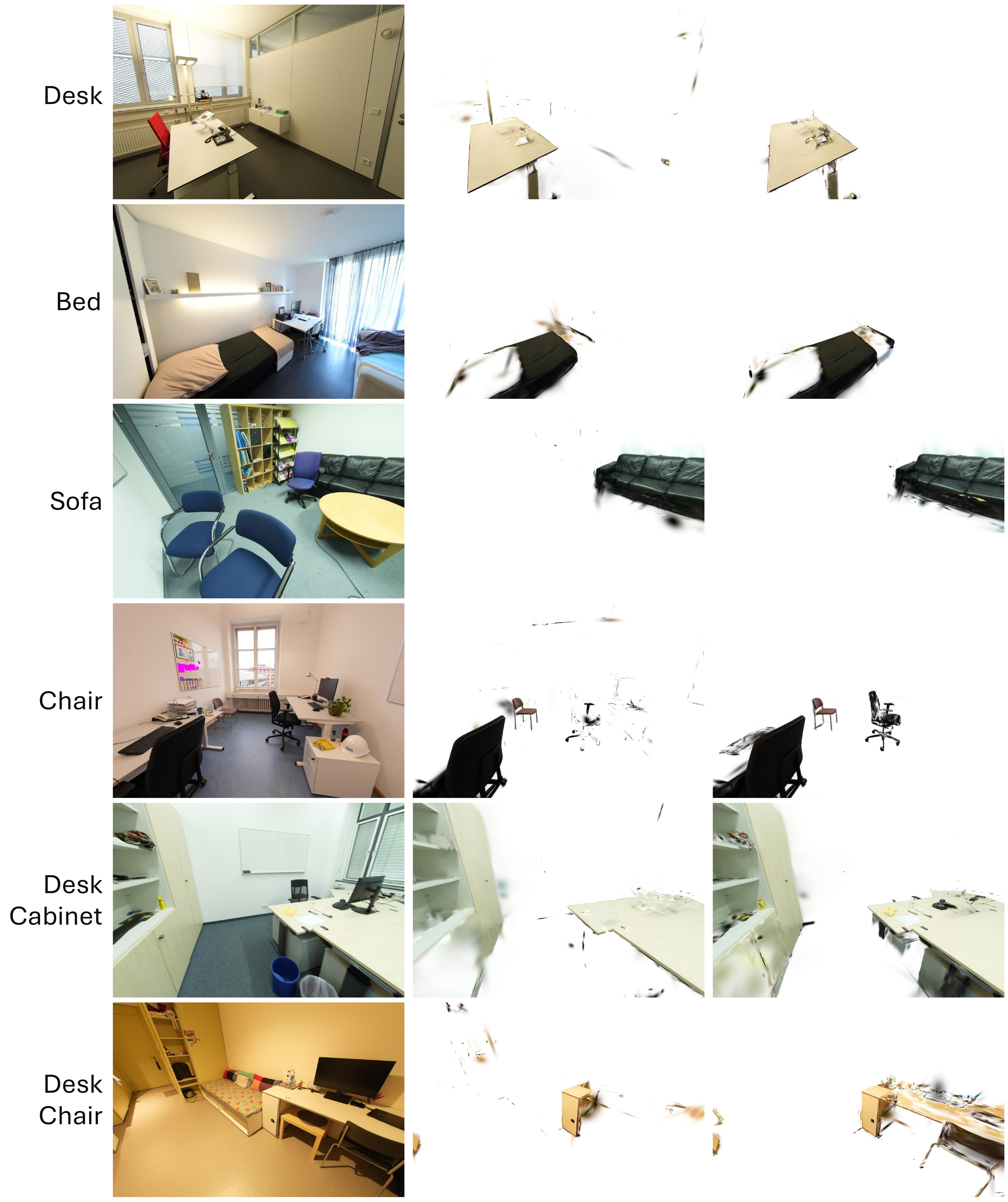}
    \caption{Example of object-level segmentation visualization results. Our method produces cleaner object-centric renderings with fewer fragmented or noisy regions across diverse object categories and multi-object queries.
    }
    \label{fig:sup_qualitative}
\end{figure}

Fig.~\ref{fig:sup_qualitative2} and Fig.~\ref{fig:sup_qualitative} show object-level segmentation visualizations obtained by activating only the Gaussian groups associated with target categories. From top to bottom, we visualize desk, bed, sofa, chair, desk+cabinet, and desk+chair. Compared with the baseline, our method produces cleaner object-level renderings with fewer noisy or unrelated Gaussian regions.

These results indicate that our method organizes 3DGS features into more consistent object-centric groups and improves the grouping of Gaussians themselves, rather than only producing cleaner 2D semantic renderings. This property can be beneficial for downstream scene understanding tasks, such as Gaussian extraction and object-level editing.

\paragraph{Qualitative results on point-level 3DGS semantic segmentation}

\begin{figure}
    \centering
    \includegraphics[width=1\textwidth]{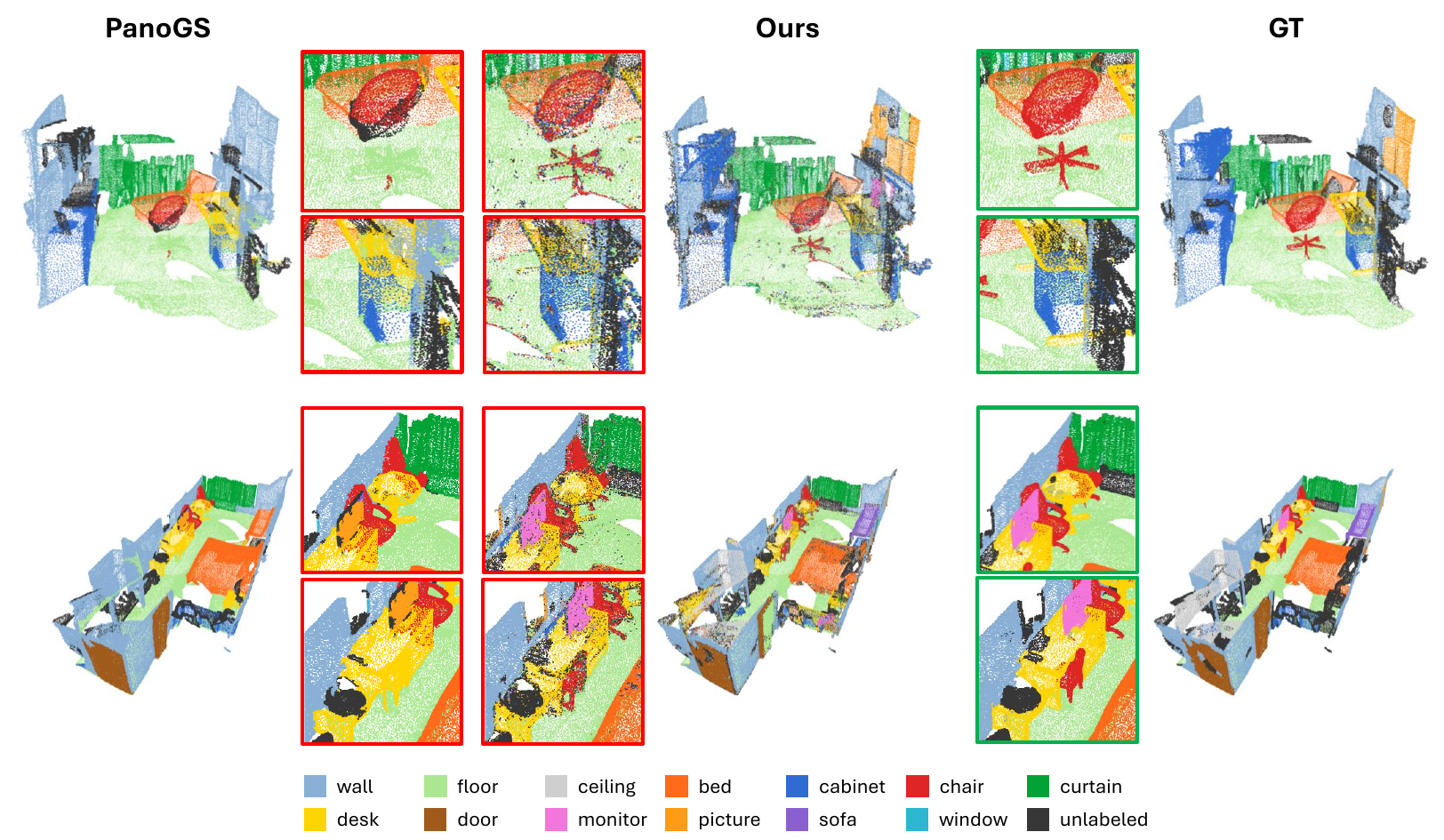}
    \caption{Detailed views of 3D point-level segmentation results on ScannetV2 from 'scene0070' and 'scene0590'. Object regions are presented to provide a clearer comparison of object-level segmentation quality.
    }
    \label{fig:sup_3dpoint}
\end{figure}

Fig.~\ref{fig:sup_3dpoint} provides close-up views of selected object regions in 3D scenes.
These visualizations demonstrate the effectiveness of our refinement strategy for 3D point cloud segmentation.
In the upper scene, our method produces a cleaner segmentation of the red chair.
By performing intra-node refinement, it filters noisy points and incorrectly assigned regions within the object group, reducing mixed-class predictions inside the object and improving object-level separation. Furthermore, inter-node refinement helps recover object parts that are weakly observed by 2D masks.
For example, the chair legs are partially assigned to background Gaussian groups in the initial representation, but our method successfully incorporates them by leveraging language-prior information during Gaussian semantic representation learning.
As a result, the final segmentation better preserves the structural completeness of the object.

A similar trend is observed in the lower scene.
Our method more clearly separates neighboring objects within the same space while preserving fine-grained structures such as thin chair legs.
These results indicate that the proposed refinement process improves both local semantic consistency and object-level structural completeness.

\end{document}